\documentclass[a4paper,11pt]{article}

\usepackage{fourier}
\usepackage{color}
\usepackage{graphicx}
\usepackage{url}
\usepackage[affil-it]{authblk}
\usepackage{amsmath}
\usepackage{wrapfig}
\usepackage[T1]{fontenc}
\usepackage{times}

\begin{document}

\title{SCULPT: Training Edge Vision Models for Post-Training Quantization Readiness}

\author{Bharadwaj Kavuri, Sourav Babu-PK, Varadhraj Ellapan, Pullarao Maddu, Prasad Deshpande}
\affil{Valeo Vision Systems, Tuam, Ireland 

Valeo India Pvt. Ltd, Chennai, India}
\date{}
\maketitle
\thispagestyle{empty}

\begin{abstract}
Edge vision models are difficult to deploy on resource-constrained hardware, making low-bit post-training quantization (PTQ) attractive. In practice, standard FP32 training often produces heavy-tailed activation distributions whose outliers destabilize activation quantization: preserving the full range wastes quantization bins on rare extremes, while aggressive clipping causes information loss. Existing solutions typically rely on quantization-aware training (QAT), which adds training complexity and bit-width coupling, or advanced PTQ procedures that repair the model after training.

We present SCULPT (Statistical Clipping and Uniform Loss for Post-Training), a training-time method that improves PTQ readiness during ordinary FP32 fine-tuning. SCULPT combines a topology-aware activation regularizer that suppresses quantization-hostile skewness and kurtosis with a stable percentile-based clipping mechanism that learns deployment-ready activation bounds. Unlike QAT, SCULPT does not simulate quantization during optimization; unlike post hoc outlier-repair PTQ methods, it does not require runtime activation transformations. The learned clipping bounds can be exported directly into a standard PTQ workflow for low-bit deployment, including INT8 and lower-bit settings such as W4A8.
\end{abstract}

\textbf{Keywords:} Post-training quantization, Activation outliers, Edge vision, Model compression

\section{Introduction}

Edge vision and perception models remain difficult to deploy on resource-constrained hardware due to tight latency, memory, and power budgets. Low-bit quantization is attractive because it enables efficient integer inference while preserving existing model architectures and deployment pipelines. In practice, however, post-training quantization (PTQ) often degrades sharply at low precision when activations exhibit heavy-tailed or highly asymmetric distributions. If the quantizer preserves rare outliers, the effective resolution for the dense bulk of the distribution becomes too coarse; if it clips aggressively, those rare extremes are truncated \cite{banner2019aciq}.

\begin{figure}[t]
    \centering
    \includegraphics[width=\linewidth]{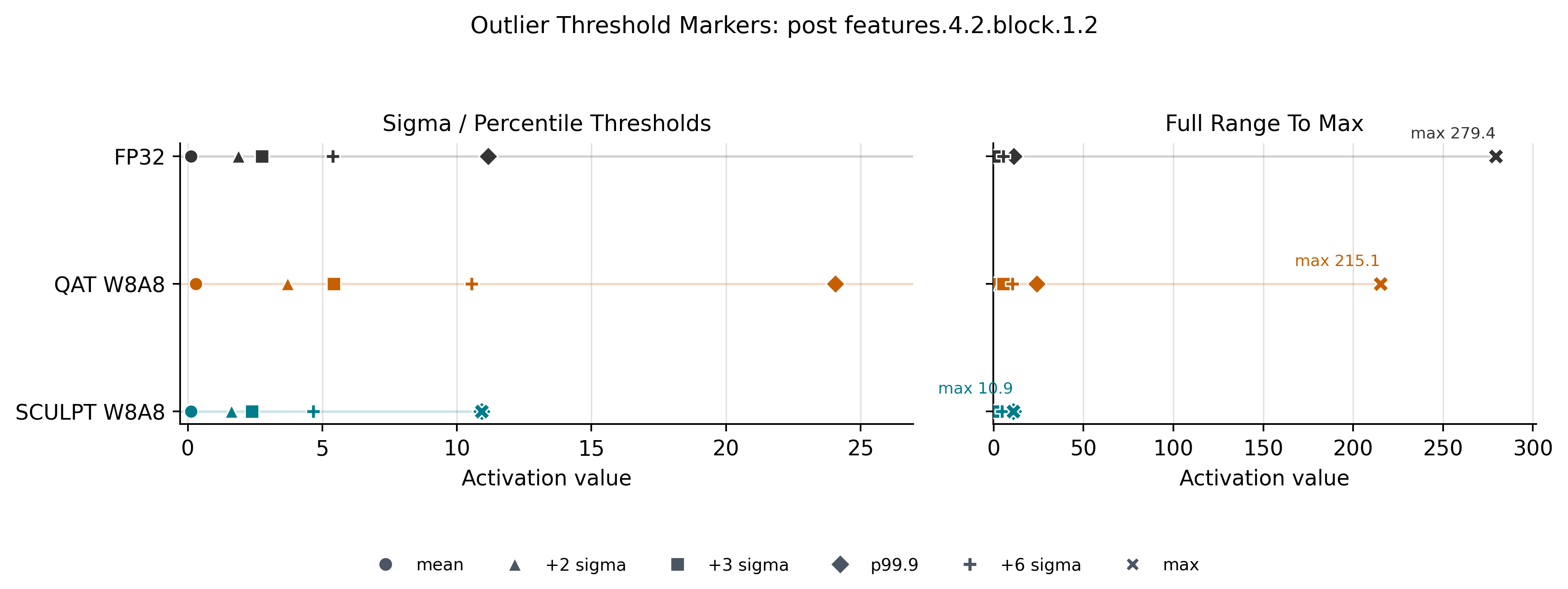}
    \caption{Outlier-threshold marker comparison for a representative hard layer (\texttt{post.features.4.2.block.1.2}). Rare extremes drive the calibration maximum far outward in standard FP32 PTQ and, to a lesser extent, in QAT. SCULPT keeps the maximum much closer to the normal layer scale, reducing outlier sensitivity during PTQ calibration.}
    \label{fig:threshold_markers_layer}
\end{figure}

Figure~\ref{fig:threshold_markers_layer} illustrates this failure mode on a representative hard layer. Most activations remain concentrated in a narrow operating region, but a small number of extreme values drive the maximum far beyond the normal scale of the layer. This mismatch makes min/max calibration brittle and is a major source of PTQ error.

Existing methods typically address this problem either during quantized training or after the full-precision model has already been trained. Quantization-aware training (QAT) and related approaches learn clipping thresholds or quantizer parameters during optimization, but they add training complexity and are often tied to a specific quantization setting \cite{choi2018pact,jain2020tqt,esser2020lsq}. On the other hand, advanced PTQ methods improve low-bit accuracy through post hoc calibration, adaptive rounding, reconstruction, or activation smoothing after training \cite{nagel2020adaround,brecq2021,xiao2023smoothquant}. These methods can be effective, but they also add workflow complexity or transformations that are unattractive in deployment-constrained settings.

In this paper, we investigate a different point in the design space: rather than repairing quantization failure only after training, can a model be trained in full precision so that it is already easier to quantize later? Our central hypothesis is that a substantial part of PTQ brittleness originates from activation distributions produced by ordinary FP32 optimization, especially the emergence of structured activation outliers that are poorly matched to low-bit uniform quantizers. To test this idea, we propose \textbf{SCULPT} (\textbf{S}tatistical \textbf{C}lipping and \textbf{U}niform \textbf{L}oss for \textbf{P}ost-\textbf{T}raining), a training-time method for improving PTQ readiness during ordinary FP32 fine-tuning.

SCULPT combines two components. First, \emph{Uniform Activation Distribution Regularization} (UADR) penalizes excess skewness and kurtosis in selected intermediate activations. Second, \emph{StablePercentileClip} introduces stable percentile-based clipping bounds that are adaptive during training and frozen by the end of optimization, yielding activation limits that can be exported directly to a standard PTQ workflow. In contrast to QAT, SCULPT does not simulate quantization during training; in contrast to post hoc outlier-repair methods, it does not require runtime activation transformations.

Our contributions are:
\begin{itemize}
    \item We identify structured activation outliers from standard FP32 training as a major obstacle to low-bit PTQ.
    \item We propose SCULPT, combining topology-aware activation regularization with stable percentile clipping to improve PTQ readiness.
    \item We evaluate SCULPT under a matched quantization contract and analyze accuracy, efficiency, and activation outlier behavior.
\end{itemize}

\section{Related Work}

PTQ converts a trained full-precision network into a lower-precision representation, typically using affine or symmetric quantizers \cite{jacob2018quantization}. Its accuracy depends heavily on how well the learned activation distributions match the target quantizer. For a uniform quantizer with clipping range $[-\alpha,\alpha]$ or $[0,\alpha]$, increasing $\alpha$ preserves outliers but coarsens the quantization grid, while decreasing $\alpha$ improves resolution for the dense central mass but clips the tails \cite{banner2019aciq}. This tradeoff becomes especially severe at low bit-widths.

There is a large body of prior work on improving quantization robustness during training. PACT learns activation clipping parameters \cite{choi2018pact}, TQT learns quantization thresholds through backpropagation \cite{jain2020tqt}, LSQ learns step sizes for low-bit training \cite{esser2020lsq}, and Robust Quantization studies training models resilient across quantization settings \cite{shkolnik2020robust}. These methods typically optimize under simulated quantization and are therefore naturally viewed as QAT-style approaches.

Another line of work improves PTQ after training through post hoc correction. AdaRound adapts rounding decisions \cite{nagel2020adaround}, BRECQ performs block-wise reconstruction \cite{brecq2021}, and SmoothQuant reduces activation outliers by shifting difficulty from activations to weights through an equivalent offline transformation \cite{xiao2023smoothquant}. Recent vision PTQ work has also shown that activation distributions after nonlinearities can be poorly matched to standard uniform quantizers, motivating specialized activation quantization schemes such as PTQ4ViT \cite{yuan2022ptq4vit}. SCULPT targets a different operating point: instead of simulating quantization during training or repairing the model after training, it modifies the FP32 fine-tuning trajectory so that the resulting model is more compatible with a later PTQ step.

\section{Method}

We propose \textbf{SCULPT}, a training-time method for improving the post-training quantization (PTQ) readiness of neural networks during ordinary FP32 fine-tuning. SCULPT combines two complementary components: (1) a topology-aware activation regularizer that suppresses quantization-hostile skewness and heavy tails, and (2) a stable percentile-based clipping layer that learns deployment-ready activation bounds. The regularizer provides soft distribution shaping, while the clipping layer provides hard boundary enforcement.

\begin{figure}[t]
    \centering
    \includegraphics[width=0.92\linewidth]{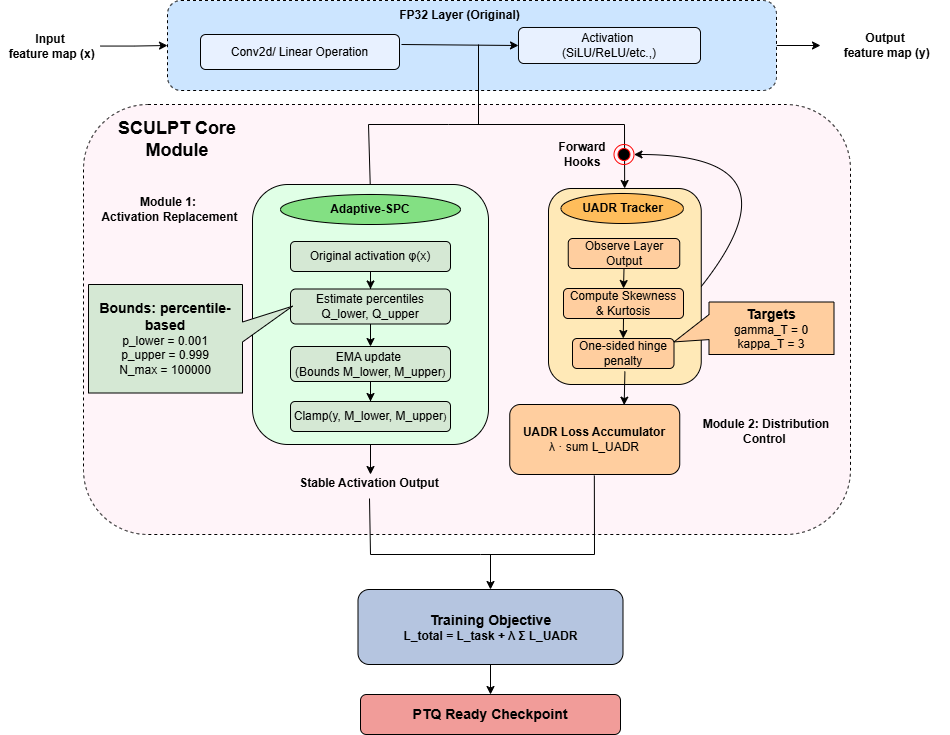}
    \caption{Overview of SCULPT. Adaptive-SPC learns stable activation bounds for PTQ export, while UADR penalizes excess skewness and kurtosis.}
    \label{fig:sculpt_overview}
\end{figure}

Figure~\ref{fig:sculpt_overview} summarizes the full SCULPT pipeline. During fine-tuning, Adaptive Stable Percentile Clipping (Adaptive-SPC) wraps standard activations with percentile-based clamped activations whose bounds are stabilized over time and frozen for deployment. In parallel, UADR tracks activation statistics from selected intermediate tensors and penalizes excess skewness and kurtosis using topology-aware one-sided targets.

\subsection{Problem Setup}

Let $f_{\theta}$ denote a model trained in full precision on task loss $\mathcal{L}_{\text{task}}$. Our goal is not to quantize the model during training, but to modify the FP32 training trajectory so that the final model is more robust to later PTQ at low precision. We focus on intermediate activations, which are often a dominant source of PTQ error when their dynamic ranges are governed by rare but large outliers. SCULPT seeks to reduce this tradeoff at its source by shaping activation statistics during FP32 fine-tuning.

\subsection{StablePercentileClip}

StablePercentileClip learns stable activation bounds during training and exports them directly for later PTQ. Unlike fixed clipping functions such as ReLU6, the clipping range is estimated from observed activation statistics and refined throughout fine-tuning. Given an input tensor $X^{(t)}$ at step $t$, we first apply the wrapped activation function:
\begin{equation}
A^{(t)} = \phi\!\left(X^{(t)}\right),
\end{equation}
where $\phi(\cdot)$ denotes the original activation (e.g., SiLU, ReLU, ReLU6, or Hardswish). Adaptive-SPC estimates percentile bounds on the output of the wrapped activation, not on the pre-activation tensor. StablePercentileClip therefore wraps the original activation rather than replacing it, preserving the underlying nonlinearity while adding learned post-activation bounds.

Let $V^{(t)} \in \mathbb{R}^{N}$ denote the flattened version of $A^{(t)}$. To control sorting cost, we form a strided subsample $\widetilde{V}^{(t)}$ using
\begin{equation}
s = \max\left(1, \left\lfloor \frac{N}{N_{\max}} \right\rfloor \right),
\end{equation}
and
\begin{equation}
\widetilde{V}^{(t)} = \left\{ V^{(t)}_{i \cdot s} \right\}_{i=0}^{\lfloor N/s \rfloor - 1}.
\end{equation}

We then compute detached lower and upper percentile estimates:
\begin{equation}
Q^{(t)}_{\text{lower}} = \operatorname{Quantile}\left(\widetilde{V}^{(t)}, p_{\text{lower}}\right), \qquad
Q^{(t)}_{\text{upper}} = \operatorname{Quantile}\left(\widetilde{V}^{(t)}, p_{\text{upper}}\right),
\end{equation}
\begin{equation}
\widehat{Q}^{(t)}_{\text{lower}} = \operatorname{sg}\!\left(Q^{(t)}_{\text{lower}}\right), \qquad
\widehat{Q}^{(t)}_{\text{upper}} = \operatorname{sg}\!\left(Q^{(t)}_{\text{upper}}\right).
\end{equation}

We maintain running lower and upper bounds updated with inverse-time momentum:
\begin{equation}
\alpha_t = \frac{\alpha_0}{1 + \gamma t},
\end{equation}
\begin{equation}
M^{(t)}_{\text{lower}} = (1-\alpha_t) M^{(t-1)}_{\text{lower}} + \alpha_t \widehat{Q}^{(t)}_{\text{lower}},
\end{equation}
\begin{equation}
M^{(t)}_{\text{upper}} = (1-\alpha_t) M^{(t-1)}_{\text{upper}} + \alpha_t \widehat{Q}^{(t)}_{\text{upper}}.
\end{equation}

The clipped activation is then
\begin{equation}
Y^{(t)} = \operatorname{clip}\!\left(A^{(t)}, M^{(t)}_{\text{lower}}, M^{(t)}_{\text{upper}}\right).
\end{equation}

Because the running bounds are updated through detached statistics, gradients do not propagate through the quantile computation itself. During backpropagation, the layer behaves like a hard clamp applied after the wrapped activation.

\subsection{UADR}

Stable clipping constrains activation range but does not explicitly discourage skewed or heavy-tailed activation shapes. To complement it, we apply \textbf{UADR} to selected learnable-layer outputs. Here, ``topology-aware'' refers to the functional position of an activation within a computation block, such as pre-activation linear outputs versus post-activation tensors. Although we retain the acronym UADR, the regularizer does not force activations toward a strictly uniform distribution; instead, it suppresses excess skewness and heavy tails using topology-aware one-sided targets.

Let $X^{(l)} \in \mathbb{R}^{N}$ denote the flattened output tensor of learnable layer $l$. We compute the empirical mean and standard deviation:
\begin{equation}
\mu_l = \frac{1}{N} \sum_{i=1}^{N} X^{(l)}_i, \qquad
\sigma_l = \sqrt{\frac{1}{N} \sum_{i=1}^{N} \left(X^{(l)}_i - \mu_l\right)^2 + \epsilon},
\end{equation}
and standardized activations
\begin{equation}
Z^{(l)}_i = \frac{X^{(l)}_i - \mu_l}{\sigma_l}.
\end{equation}

Skewness and kurtosis are then
\begin{equation}
\gamma\!\left(X^{(l)}\right) = \frac{1}{N} \sum_{i=1}^{N} \left(Z^{(l)}_i\right)^3, \qquad
\kappa\!\left(X^{(l)}\right) = \frac{1}{N} \sum_{i=1}^{N} \left(Z^{(l)}_i\right)^4.
\end{equation}

In the experiments of this paper, we use the simpler setting
\begin{equation}
\gamma_T^{(l)} = 0
\end{equation}
for all regularized layers, and a shared upper target $\kappa_T = 3$.

The per-layer UADR loss is
\begin{equation}
\mathcal{L}_{\text{UADR}}^{(l)} =
\alpha \, \max\!\left(0, \gamma\!\left(X^{(l)}\right) - \gamma_T^{(l)}\right)^2
+
\beta \, \max\!\left(0, \kappa\!\left(X^{(l)}\right) - \kappa_T\right)^2.
\end{equation}

Let $\mathcal{S}$ denote the set of layers to which UADR is applied. In our implementation, UADR is applied to Conv2d and Linear outputs, while Adaptive-SPC is inserted after SiLU, ReLU, ReLU6, and Hardswish activations. The total training objective is
\begin{equation}
\mathcal{L}_{\text{total}} =
\mathcal{L}_{\text{task}} +
\lambda \sum_{l \in \mathcal{S}} \mathcal{L}_{\text{UADR}}^{(l)}.
\end{equation}

UADR and StablePercentileClip play complementary roles: UADR reduces the tendency of intermediate activations to develop excessive skewness and heavy tails, while StablePercentileClip enforces explicit activation bounds that stabilize into deployment-time clipping limits. After fine-tuning, the learned bounds are frozen and exported as layerwise activation limits for deployment, so the resulting model can be quantized using a standard PTQ workflow without simulated quantization during training and without runtime outlier-repair transforms.

\section{Experimental Setup}

We evaluate SCULPT on EfficientNet-B0 \cite{tan2019efficientnet} with ImageNette \cite{howard2019imagenette, deng2009imagenet} as a controlled low-bit quantization testbed. The baseline model is initialized from Torchvision ImageNet-1K EfficientNet-B0 weights, adapted to 10 classes, and trained in FP32 for 80 epochs using AdamW, label smoothing of 0.1, RandAugment, automatic mixed precision (AMP), batch size 128, and a cosine OneCycle learning-rate schedule.

Starting from the trained FP32 checkpoint, SCULPT fine-tunes the model with Adaptive Stable Percentile Clipping inserted after all SiLU, ReLU, ReLU6, and Hardswish activations, and with UADR regularization applied to all Conv2d and Linear outputs. Adaptive-SPC estimates lower and upper activation percentile bounds at 0.001 and 0.999, respectively, using an inverse-time exponential moving average. We set $N_{\max}=100000$ for percentile estimation in all experiments. These bounds are frozen after epoch 2 and converted into static activation-plus-clamp modules for deployment-time PTQ.

For UADR, we use a one-sided hinge penalty with target skewness $\gamma_T = 0$, target kurtosis $\kappa_T = 3$, $\alpha = 0.01$, $\beta = 0.01$, and global regularization weight $\lambda = 5\times10^{-4}$, which is reduced by a factor of 0.1 after SPC freezing.

We compare standard FP32 PTQ, SCULPT PTQ, and QAT under an identical supergroup QDQ contract. The PTQ setup uses per-channel symmetric weight quantization, per-tensor asymmetric activation quantization, and min/max calibration over 32 calibration batches. QDQ nodes are placed before Conv/Linear modules and after activation or SPC clamp modules, with no QDQ inserted between Conv and SiLU/ReLU/SPC. The QAT baseline uses native PyTorch fake quantizers under the same supergroup boundaries.

Checkpoint selection is based on quantized validation accuracy under the matched supergroup PTQ contract rather than FP32 validation accuracy. For FP32, SCULPT, and QAT runs, epoch-level PTQ tracking is used to identify the best deployment-relevant checkpoint.

\section{Experimental Results}

\subsection{Main Quantization Results}

Table~\ref{tab:main_results} compares standard FP32 PTQ, SCULPT-conditioned PTQ, and QAT baselines under the same supergroup QDQ contract. SCULPT substantially improves low-bit PTQ relative to naive FP32 PTQ. In the primary W8A8 setting, standard FP32 PTQ achieves 90.5478\% top-1 accuracy, whereas SCULPT PTQ reaches 99.3631\%, an absolute gain of 8.8153 percentage points. Under the same W8A8 setting, the tested QAT baseline reaches 98.2420\%, which remains below SCULPT PTQ by 1.1211 points.

The same trend holds at W4A8. Standard FP32 PTQ drops sharply to 40.5605\%, whereas SCULPT PTQ reaches 80.2803\%, slightly exceeding the tested QAT W8A8 PTQ baseline at 79.4395\%. These results suggest that conditioning activation statistics during FP32 fine-tuning can substantially improve downstream PTQ robustness, even without simulated quantization during training.

\begin{table}[!h]
\begin{center}
\begin{tabular}{|l|c|c|c|c|}
\hline
Method & FP32 & W8A8 PTQ & W4A8 PTQ & Best PTQ Step \\
\hline
FP32      & 99.2866 & 90.5478 & 40.5605 & 3577 \\
\hline
SCULPT    & 99.5159 & 99.3631 & 80.2803 & 584 \\
\hline
QAT W8A8  & 97.7325 & 98.2420 & 79.4395 & 2044 \\
\hline
QAT W4A8  & 93.1465 & 89.9618 & 98.2675 & 2044 \\
\hline
\end{tabular}
\end{center}
\vspace{-8pt}
\caption{Main results under the shared QDQ contract.}
\label{tab:main_results}
\vspace{-8pt}
\end{table}

SCULPT reaches its best W8A8 PTQ result after 584 optimizer steps, compared with 2044 for the tested W8A8 QAT baseline and 3577 for naive FP32 PTQ. Its full run is also shorter: 16 epochs / 1168 steps, versus 30 epochs / 2190 steps for both QAT baselines and 80 epochs / 5840 steps for the original FP32 run.

Although dedicated low-bit QAT remains stronger in its target setting, SCULPT is designed to improve PTQ readiness without simulated quantization during training, and is therefore most directly compared against naive PTQ and the matched W8A8 QAT baseline in the primary deployment setting.

\subsection{Activation Distribution Analysis}

To understand why SCULPT improves PTQ accuracy, we analyze the activation distributions observed after training. Across the 131 encoded activation boundaries used in the quantized model, our layer-wise analysis shows that SCULPT reduces the mean PTQ NRMSE/std from 0.04818 for standard FP32 PTQ to 0.01519, and reduces the mean absmax/std from 42.04 to 11.56. These reductions are consistent with the intended role of SCULPT: suppressing the rare extreme values that otherwise dominate min/max calibration and stretch the quantization range.

\begin{figure}[t]
    \centering
    \includegraphics[width=\linewidth]{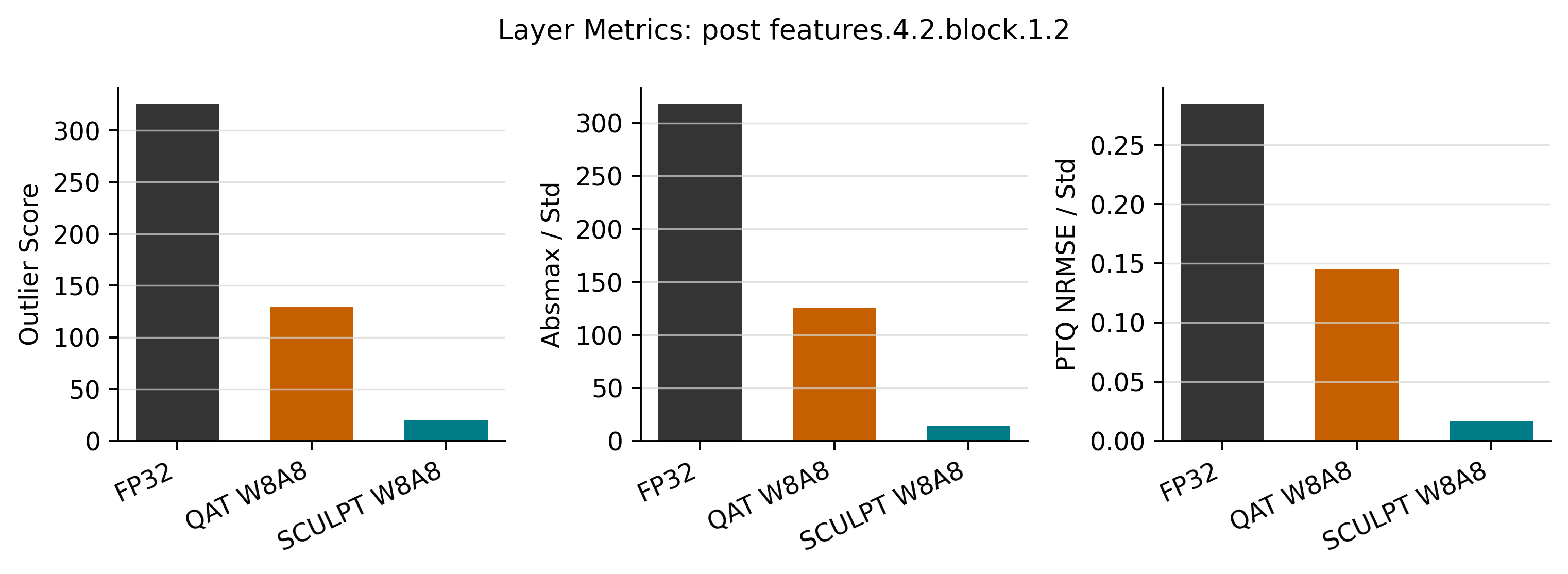}
    \caption{Scalar metrics for a representative hard layer. Lower is better in all three cases. SCULPT sharply reduces outlier score, absmax/std, and PTQ NRMSE/std relative to both standard FP32 PTQ and the tested W8A8 QAT baseline.}
    \label{fig:layer_metrics_layer}
\end{figure}

Figure~\ref{fig:layer_metrics_layer} summarizes a representative hard layer (\texttt{post.features.4.2.block.1.2}) using scalar metrics. SCULPT reduces the outlier score from 325.08 in FP32 PTQ and 129.12 in QAT to 19.99, reduces absmax/std from 317.6 and 125.8 to 14.4, and reduces normalized PTQ error from 0.2841 and 0.1452 to 0.0163, respectively. These results indicate that SCULPT does not merely shift calibration slightly; it materially changes the tail behavior that makes PTQ brittle in the first place.

\subsection{Discussion}

These results suggest that a substantial portion of PTQ brittleness can be mitigated before quantization itself, by modifying the activation statistics learned during FP32 fine-tuning. In this controlled benchmark, SCULPT yields a substantially stronger PTQ checkpoint than naive FP32 PTQ and also compares favorably to the tested W8A8 QAT baseline while requiring fewer optimization steps to reach its best quantized result.

\section{Conclusion}

We presented \textbf{SCULPT}, a training-time method for improving post-training quantization (PTQ) readiness during ordinary FP32 fine-tuning. SCULPT combines a topology-aware activation regularizer that suppresses quantization-hostile skewness and heavy tails with a stable percentile-based clipping mechanism that learns deployment-ready activation bounds. Together, these components condition the full-precision model so that later low-bit PTQ becomes substantially more robust without simulated quantization during training or runtime outlier-repair transforms.

On EfficientNet-B0/ImageNette under an identical supergroup QDQ contract, SCULPT substantially outperforms naive FP32 PTQ and compares favorably to the tested W8A8 QAT baseline. In the primary W8A8 setting, SCULPT PTQ reaches 99.3631\% top-1 accuracy, compared with 90.5478\% for standard FP32 PTQ and 98.2420\% for the tested W8A8 QAT baseline. These results support the hypothesis that a substantial portion of PTQ brittleness can be mitigated before quantization itself. Extending the same analysis to more complex perception models remains an important next step.

\section*{Acknowledgments}
Generative AI assisted with language refinement and structural editing; the authors verified all technical claims and remain responsible for the manuscript. The authors gratefully acknowledge Valeo Vision Systems, Tuam, Ireland, for providing support and resources for this work.

\bibliographystyle{apalike}
\bibliography{imvip}

@inproceedings{jacob2018quantization,
  author = {Jacob, Benoit and Kligys, Skirmantas and Chen, Bo and Zhu, Menglong and Tang, Matthew and Howard, Andrew and Adam, Hartwig and Kalenichenko, Dmitry},
  title = {Quantization and Training of Neural Networks for Efficient Integer-Arithmetic-Only Inference},
  booktitle = {Proceedings of the IEEE Conference on Computer Vision and Pattern Recognition (CVPR)},
  year = {2018}
}

@article{choi2018pact,
  author = {Choi, Jungwook and Wang, Zhuo and Venkataramani, Swagath and Chuang, Pierce I-Jen and Srinivasan, Vijayalakshmi and Gopalakrishnan, Kailash},
  title = {{PACT}: Parameterized Clipping Activation for Quantized Neural Networks},
  journal = {arXiv preprint arXiv:1805.06085},
  year = {2018}
}

@inproceedings{jain2020tqt,
  author = {Jain, Sambhav and Gural, Albert and Wu, Michael and Dick, Chris},
  title = {Trained Quantization Thresholds for Accurate and Efficient Fixed-Point Inference of Deep Neural Networks},
  booktitle = {Proceedings of Machine Learning and Systems (MLSys)},
  year = {2020}
}

@inproceedings{esser2020lsq,
  author = {Esser, Steven K. and McKinstry, Jeffrey L. and Bablani, Deepika and Appuswamy, Rathinakumar and Modha, Dharmendra S.},
  title = {{Learned Step Size Quantization}},
  booktitle = {International Conference on Learning Representations (ICLR)},
  year = {2020}
}

@inproceedings{banner2019aciq,
  author = {Banner, Ron and Nahshan, Yury and Hoffer, Elad and Soudry, Daniel},
  title = {{ACIQ}: Analytical Clipping for Integer Quantization of Neural Networks},
  booktitle = {International Conference on Learning Representations (ICLR)},
  year = {2019}
}

@inproceedings{nagel2020adaround,
  author = {Nagel, Markus and Amjad, Rana Ali and Van Baalen, Mart and Louizos, Christos and Blankevoort, Tijmen},
  title = {Up or Down? Adaptive Rounding for Post-Training Quantization},
  booktitle = {Proceedings of the 37th International Conference on Machine Learning (ICML)},
  year = {2020}
}

@inproceedings{brecq2021,
  author = {Li, Yuhang and Gong, Ruihao and Tan, Xu and Yang, Yang and Hu, Peng and Zhang, Qi and Yu, Fengwei and Wang, Wei and Gu, Shi},
  title = {{BRECQ}: Pushing the Limit of Post-Training Quantization by Block Reconstruction},
  booktitle = {International Conference on Learning Representations (ICLR)},
  year = {2021}
}

@inproceedings{shkolnik2020robust,
  author = {Shkolnik, Moran and Chmiel, Brian and Banner, Ron and Shomron, Gil and Nahshan, Yury and Bronstein, Alex and Weiser, Uri},
  title = {Robust Quantization: One Model to Rule Them All},
  booktitle = {Advances in Neural Information Processing Systems (NeurIPS)},
  year = {2020}
}

@inproceedings{xiao2023smoothquant,
  author = {Xiao, Guangxuan and Lin, Ji and Seznec, Mickael and Wu, Hao and Demouth, Julien and Han, Song},
  title = {SmoothQuant: Accurate and Efficient Post-Training Quantization for Large Language Models},
  booktitle = {Proceedings of the 40th International Conference on Machine Learning (ICML)},
  year = {2023}
}

@inproceedings{yuan2022ptq4vit,
  author = {Yuan, Zhihang and Xue, Chenhao and Chen, Yiqi and Wu, Qiang and Sun, Guangyu},
  title = {{PTQ4ViT}: Post-Training Quantization for Vision Transformers with Twin Uniform Quantization},
  booktitle = {European Conference on Computer Vision (ECCV)},
  year = {2022}
}

@inproceedings{tan2019efficientnet,
  title={Efficientnet: Rethinking model scaling for convolutional neural networks},
  author={Tan, Mingxing and Le, Quoc},
  booktitle={International conference on machine learning},
  pages={6105--6114},
  year={2019},
  organization={PMLR}
}

@misc{howard2019imagenette,
  title={Imagenette: A smaller subset of 10 easily classified classes from ImageNet},
  author={Howard, Jeremy},
  year={2019},
  howpublished={\url{https://github.com/fastai/imagenette}}
}

@inproceedings{deng2009imagenet,
  title={ImageNet: A Large-Scale Hierarchical Image Database},
  author={Deng, Jia and Dong, Wei and Socher, Richard and Li, Li-Jia and Li, Kai and Fei-Fei, Li},
  booktitle={2009 IEEE Conference on Computer Vision and Pattern Recognition},
  pages={248--255},
  year={2009},
  organization={IEEE}
}

\end{document}